# AliMe KBQA: Question Answering over Structured Knowledge for E-commerce Customer Service

Feng-Lin Li, Weijia Chen, Qi Huang, Yikun Guo

Alibaba Group.
Hangzhou, China, 311100
{fenglin.lfl, yikun.gyk}@alibaba-inc.com

**Abstract.** With the rise of knowledge graph (KG), question answering over knowledge base (KBQA) has attracted increasing attention in recent years. Despite much research has been conducted on this topic, it is still challenging to apply KBQA technology in industry because business knowledge and real-world questions can be rather complicated. In this paper, we present AliMe-KBQA, a bold attempt to apply KBQA in the E-commerce customer service field. To handle real knowledge and questions, we extend the classic "subject-predicate-object (SPO)" structure with property hierarchy, key-value structure and compound value type (CVT), and enhance traditional KBQA with constraints recognition and reasoning ability. We launch AliMe-KBQA in the *Marketing Promotion* scenario for merchants during the "Double 11" period in 2018 and other such promotional events afterwards. Online results suggest that AliMe-KBQA is not only able to gain better resolution and improve customer satisfaction, but also becomes the preferred knowledge management method by business knowledge staffs since it offers a more convenient and efficient management experience.

## 1 Introduction

**AliMe** [13] is an intelligent assistant that offers *after-sales service* in the E-commerce customer service field. With question-answer pair (QA [1]) knowledge representation and deep learning (DL) based text matching, AliMe has achieved remarkable success in the consumer community: it currently serves millions of customer questions per day and is able to address 90%+ of them. However, there is still room for improvement in the merchant community, where customer questions are more scattered and complicated.

In our observation, there are still several challenges in knowledge management and for AliMe to better understand customer questions. First, our knowledge is organized as QA pairs, which are widely used as index for knowledge, rather than true knowledge. The treatment of **knowledge as QA pairs** introduces redundancy (a piece of knowledge has to be enumerated as to deal with real life scenarios, for example, the two QA pairs "一个账号可以绑定几个手机号？how many phone numbers can be bound to a Taobao account" and "为什么我的账号绑定手机时说超出限制？why the system shows that the number of phone numbers exceeds the limit when binding a phone to my account?" actually refers to the same piece of knowledge "一个账号可以绑定 6 个手机号码 a Taobao account can associate with at most 6 phone numbers"), not to mention sometimes questions can not be exhaustively listed, especially for instance-level and compositional knowledge (for example, knowledge staffs have to maintain such a question "how to register for a promotion program" for at least dozens of programs in AliMe).

---

[1] We use "QA" to refer to "question answering" and "question-answer" interchangeably.

Second, with QA representation, business knowledge staffs have to constantly analyse regulations and elicit frequently asked questions (FAQs) for similar and even repeated scenarios like promotional programs, which can be largely alleviated by defining a common schema structure for customer questions. Third, QA pairs do not support reasoning, which is indeed needed in customer service where regulations are of key importance. For example, we have a regulation "店铺级优惠和店铺级优惠不能同时使用 (In-store and in-store discount can not be applied at the same time)", but do not index a specific instance-level knowledge like "优惠券和店铺红包不能同时使用 (Coupon and in-store red packet can not used at the same time)", where "优惠券 (Coupon)" and "店铺红包 (In-store Red Packet)" are both of type "In-store Discount".

To address these challenges, we launched **Knowledge Cloud** project, aiming at constructing a systematically structured knowledge representation and enabling AliMe to better understand customer questions instead of simply matching questions to knowledge items based on text or semantic similarity.

In this paper, we present AliMe-KBQA, a knowledge graph based bot application, and introduce its underlying knowledge representation and supporting techniques. To the best of our knowledge, this is the first attempt to apply KBQA techniques in customer service industry on a large scale. Our paper makes the following contributions:

– Extends the classic SPO structure to capture practical knowledge and questions: (1) we use property hierarchy instead of flatten properties to organize knowledge, and guide vague questions; (2) we adopt key-value structure and Compound Value Type (CVT) to characterize complicated answer for precise question answering.
– Empower traditional KBQA with constraints recognition and reasoning ability based on our structured knowledge representation in support of complicated and regulation-oriented QA in the E-commerce customer service field.
– Launch AliMe-KBQA as a practical QA bot in the *Marketing Promotion* scenario. Online results suggestion that KBQA is able to not only gain better resolution rate and degree of satisfaction from customer side, but also be preferred by business knowledge staffs since it offers better knowledge management experience.

The rest of the paper is structured as follows: Section 2 presents our structured knowledge representation; Section 3 discusses the extended KBQA approach; Section 4 demonstrates system features; Section 5 reviews related work, Section 6 concludes the paper and sketches directions for future work.

## 2   Knowledge Representation

In general, a knowledge graph $\mathcal{K}$ is organzied in terms of nodes and links, and defined as a set of triples $(e_h, p, e_t) \in \mathcal{E} \times \mathcal{P} \times \mathcal{E}$ (e.g., "Beijing, capital_of, China"), where $\mathcal{E}$ denotes a set of entities/classes/literals, and $\mathcal{P}$ denotes a set of properties.

Not surprisingly, this general triple representation is insufficient in capturing practical knowledge and questions. One can see the examples shown in Table 1. The piece of knowledge $k_1$ is very general and need to be further specified, otherwise its answer will be too complicated to maintain and read. The problem with the $k_2$ is that it includes two entities "淘抢购 (Tao Flash Sale)" and "双十一 (Double 11)", and the former modifies the latter. Similarly, the entity "Double 11" in $k_3$ is a modifier of "floor price", which is also an entity as merchants often ask what it is and how to calculate it. In $k_4$, "优惠券 (Coupon)" and "单品宝 (SKU-Bao)" are specific instances of "In-store Discount", and such combination of different kinds of discount is hard to enumerate and maintain for business knowledge staffs.

**Table 1.** Example QA-style knowledge items

| |
|---|
| $k_1$. 店铺宝优惠规则 |
| The discount regulation of Store-Bao |
| $k_2$. 怎么参加淘抢购的双十一? |
| How join in the Double 11 event of Tao Flash Sale? |
| $k_3$. 淘抢购是否计入双十一最低活动价? |
| Whether Tao Flash Sale is counted in Double 11's floor price? |
| $k_4$. 优惠券和单品宝能不能一起使用? |
| Can coupon and SKU-Bao can be used in conjunction? |

To capture real business knowledge, we present our extended knowledge representation in Fig. 1, where colored rectangles denote our extension: (1) a property can be decomposed into sub-properties; (2) value type is extended with key-value structure and Compound Value Type (CVT). In our ontology, a property is treated as a mapping function that maps an entity of type "Class" (domain) to a value that has a type "Value_Type" (range). An entity reifies a property of its class when the value of that property is configured. Our CVT is dopted from Freebase [5], treated as self-defined class and captured as table. For simplicity, we define for each CVT table a main (answer) column that will be queried, and take the other columns as conditions or constraints.

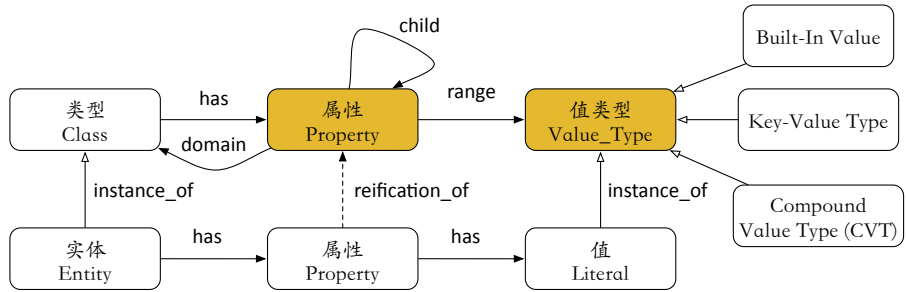

**Fig. 1.** The ontology of our structured knowledge representation

We show an excerpt of the schema of "Promotion_Tool" in Fig. 2. Nodes are entities $e \in \mathcal{E}$ (e.g., "店铺宝 Store-Bao"), classes $c \in \mathcal{C}$ (e.g., "营销工具 Promotion_Tool"), literals $l \in \mathcal{L}$ (e.g., "三星级 3-star"). Links are properteis $p \in \mathcal{P}$ (e.g., "定义 definition"). Moreover, the entity "Store-Bao" is an instance of "Promotion_Tool", and reifies its properties through associating corresponding values.

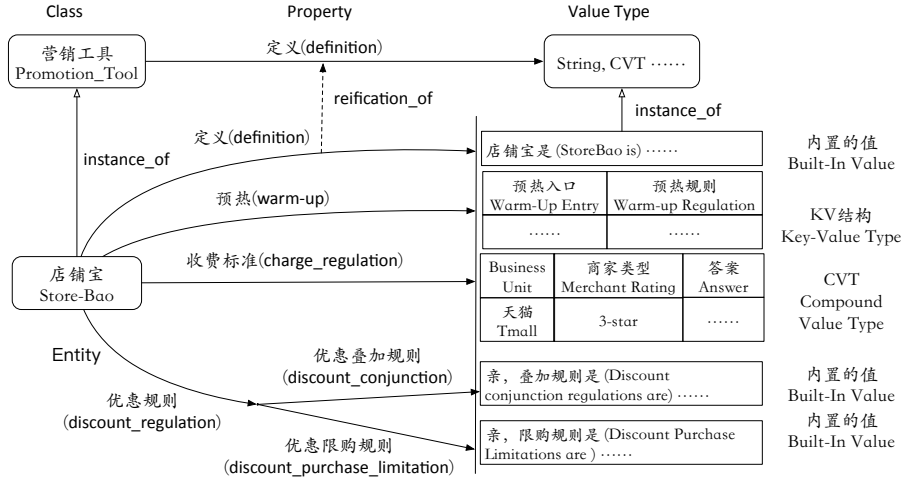

**Fig. 2.** An excerpt of our extended knowledge represenattion

A property, if composite, can be decomposed into sub-properties according to business knowledge. The resulting hierarchical structure, instead of flatten properties, offers better understandability and knowledge management experience. A path from root property to leaf property constitutes a property chain that links an entity to a corresponding value. For example, the root property "discount_regulation" in Fig. 2 is decomposed into "discount_conjunction" and "discount_purchase_limitation"; and, the path "discount_regulation - discount_conjunction" forms a property chain that maps "Store-Bao" to a text string value. An extra benefit of hierarchical property structure is that it enables a bot to guide vague [2] customer questions through recommendation. For example, if a customer merely mention "discount", we can guide customers to either entities "Store-Bao" or sub-properties "'discount_conjunction" or "discount_purchase_limitation" through recommendation according to the knowledge structure and dialog context.

The value of (leaf) property can be a simple value (captued as built-in types such as String and Integer), or a block of text (captured as String or key-value segmented wiki-style document), or a composite value (captured as CVT). The key-value structure allows to segment a long answer text on demand, is able to support tabbed UI reprsentation of the answer and bring about better reading experience. The use of CVT to characterize multiple-fields of a property, allows us to capture multi-constraints of a knowledge item (e.g., the value of "charge_regulation" depends on not only "Business Unit" but also "Merchant Rating"), and enables us to perform precise question answering (e.g., if we know the "Business Unit" and "Merchant Rating", we can get the precise answer instead of output the whole table).

---
[2] It is worth to mention that nearly a quarter of questions are vague or incomplete in practice.

Our knowledge representation is able to capture not only instance-level but also class-level knowledge (e.g., regulations). For example, in the *Marketing Promotion* scenario, we have a regulation: "店铺级优惠可以与单品级优惠和跨店级优惠叠加，但是不能和店铺级优惠相互叠加 (An in-store discount can be used in conjunction with SKU discounts and inter-store discounts, but not with other kinds of in-store discounts)". As shown in Fig. 3, we define "店铺级优惠 In-store Discount" (resp. "单品级优惠 SKU Discount" and "跨店级优惠 Inter-store Discount") as a class and design for it a property "是否可以叠加 (use_in_conjunction)". We then capture the regulation with a CVT, and link the CVT table to a cognominal entity of the class "In-store Discount" (resp. "SKU Discount" and "Inter-store Discount"). A specific discount, e.g., "优惠券 Coupon", is an instance of "In-store Discount (class)", and is also a member of the special entity "In-store Discount (entity)".

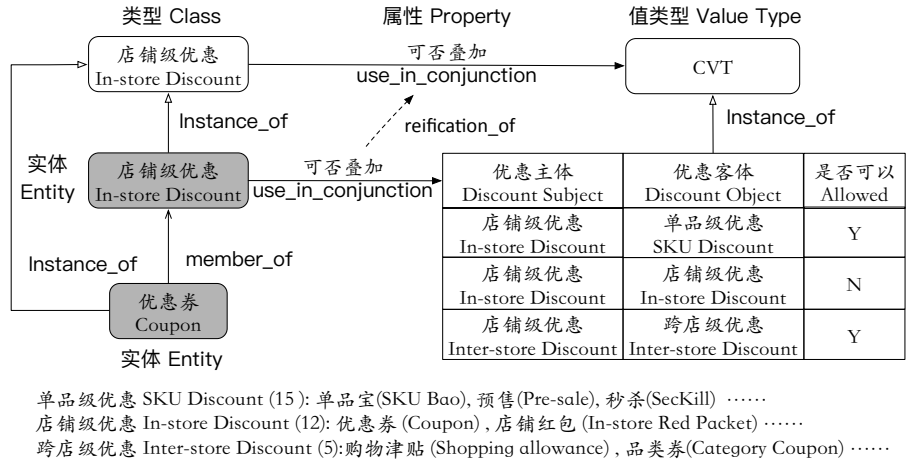

**Fig. 3.** An example of regulation knowledge

Taking a knowledge management perspective, there are 15 kinds of "SKU Discount", 12 kinds of "In-store Discount", 5 kinds of "Inter-store Discount". Using QA style representation, there would be $1024 = (15+12+5) * (15+12+5)$ QA pairs. With our structured knowledge representation, there will be only dozens of knowledge items (1 property, $9 = 3 \times 3$ regulations, $32 = 15 + 5 + 12$ "instance_of" and "member_of" tuples). That is, the structured representation largely reduces the number of knowledge items, enabling convenient knowledge management and better model matching performance (large number of similar QA pairs are difficult for model to distinguish).

With the defiend schema, given a new promotion program, business knowledge staffs only need to fill in the answer of defined properties that represent customer questions, and do not need to elicit frequently asked questions any more. More than that, the training samples for the properties of a shcema can be highly, or even totally reused (only entity mentions need to be substituted). That is, the cold-start cost can be largely reduced. In fact, it only took us one working day to launch AliMe-KBQA in the *Marketing Promotion* scenario for the "Double 12" day after its application in the "Double 11" period.

## 3   KBQA Approach

We base our KBQA approach on staged query graph generation [19] which uses knowledge graph to prune the search space, and multi-constraint query graph [2] that focuses on constraint recognition and binding [2]. We focus on how to utilize the structured knowledge representation for KBQA applications, and employ state-of-the-art DL models (CNN, Bi-GRU, attention, label embedding, etc.) to perform our task.

In this line of research, the KBQA probelm can be defined as follows: given a question $q$ and a knowledge base $\mathcal{K}$, $q$ is translated to a set of query graphs $\mathcal{G}$ according to $\mathcal{K}$, then a feature vector $f(q, g)$ is extracted for each graph $g \in \mathcal{G}$ and used for ranking, the graph with the highest score will be chosen and executed to obtain the answer $a$.

We show an overview of our approach in Fig. 4. Given a question, we at first generate a set of basic query graphs in the form of $(e, p, v)$, where $e$ denotes a topic entity, $p$ denotes a property and $v$ denotes a variable node. Specifically, we identify entities from the question through a *trie*-based rule engine [3], substitute the mentioned entities with a special symbol, and then map the masked question to candidate (leaf) properties of the classes of identified entities through a tailored CNN [10] classification model.

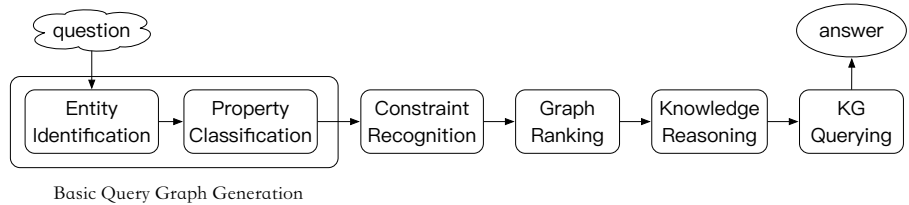

**Fig. 4.** An overview of our KBQA approach

In the case $v$ is a CVT node, a basic query graph will have the form of $(e, p, v_{cvt}, r, x)$, where $r$ denotes the answer column of the CVT table, and $x$ stands for a specific value in a CVT cell. One can refer to the bottom right corner of Fig. 5 for an example, where the entity is "Double 11" and the property is "registration_process". Further, we use rule-based and similarity-based string matching to identify constraints and link them to the CVT node. As shown at the rightmost part of Fig. 5, the recognized constraint $(Tao\_Flash\_Sale, =, y_1)$ is linked to the CVT node through $(v_{cvt}, promo\_method, y_1)$.

---
[3] Entity linking is not performed as disambiguation is not necessary in our current scenario.

Note that Fig. 5 shows only one possible query graph for the given question. As we usually have multiple query graphs, simple or complicated, we at last employ a ranking model based on LambdaRank [8] to rank candidate graphs and reply with the answer of Top-K graphs ($K = 1$ for question answering and $K = 3$ for recommendation).

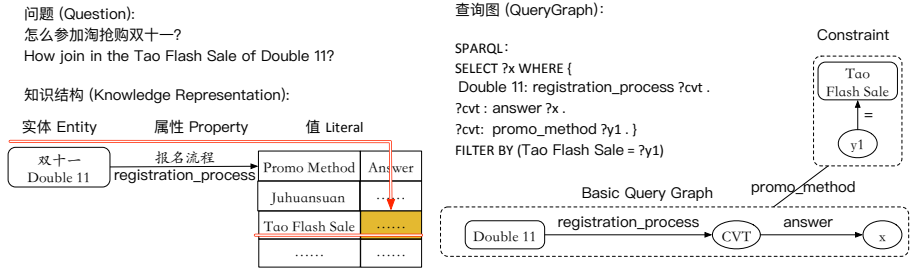

**Fig. 5.** Question answering over structured knowledge representation: an exapmle

Last but not least, we need to perform reasoning (when needed) on the ranked Top-K graphs in order to obtain the answer, especially for class-level regulations in the E-commerce customer service field. As in Fig. 6, given the question "优惠券和单品宝能不能一起使用 (Can coupon and SKU-Bao be used at the same time?)", we first form the basic graph $(e, p, v_{cvt}, r, x)$, where $e$ indicates "Coupon", $p$ stands for "use_in_conjunction", $r$ represents "answer". On observing that the domain and the range of $p$ need to be inferred [4], we generalize "Coupon" (resp. "SKU-Bao") as a kind of "In-store Discount" (resp. "SKU Discount") by following the "member_of" property shown in Fig. 3. After that, we are able to query the CVT table whether "In-store Discount" can be used together with "SKU Discount", and obtain the explicit and precise answer "NO".

## 4 KBQA System

We first launched our KBQA system in the *Marketing Promotion* scenario in AliWanxiang (a product of the AliMe family) for Tmall merchants. Our schema includes 121 properties, 73 out of which are associated with CVTs, and covers 320 original QA pairs. During the "Double 11" period in 2018, our bot served more than one million customer questions, and achieved a resolution rate [5] of 90%+, which is 10-percent higher than traditional QA-style representation and question to question text matching. Moreover, it also won a 10% increase of satisfaction degree according to our user survey.

---

[4] We use an indicator to denote whether the domain (resp. range) of a property need to be inferred (yes:1, no:0), and how it will be inferred (e.g., by following the "member_of" property).

[5] The resolution rate $rr$ is calculated as follows: $rr = 1 - U/T$, where $U$ denotes the number of unsolved sessions, which includes disliked sessions, no-answer sessions, and sessions that explicitly requests for staff service, and $T$ stands for the number of total sessions.

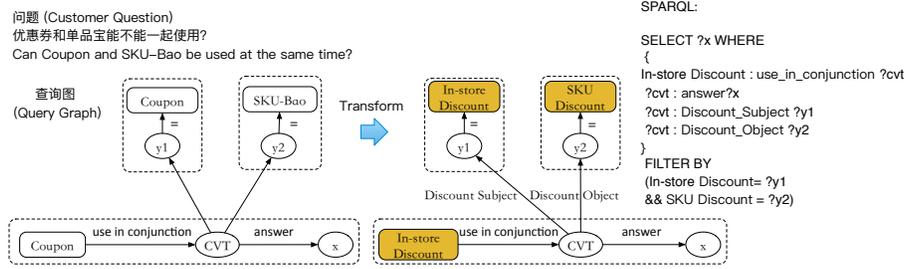

**Fig. 6.** Knowledge reasoning over structured representation: an exapmle

We further applied our structured knowledge representation and KBQA appproach in three other scenarios, and found that our representation is adequate in capturing practical knowledge and our KBQA approach is able to deal with real-world cusomer questions. The statistics in Table 2 show that the number of QA pairs can be largely reduced (we are more concerned about the compression ratio regarding QA pairs and properties $Compr_1$ becasue the number of properties indicates how many knowledge items need to label training samples for and perform model matching on), and the resolution rate gains an absolute increase of 5% on average. It also show that CVT is of importance, even essential (e.g., CVT accounts for 62.5% of the properties in scenario-3), in representing practical knowledge.

**Table 2.** Statistics of knowledge before and after structuration

| Scenarios | #QA | #Entity | #Property | $Compr_1(Compr_2)$ | #CVT | CVTr | Reslolution |
|---|---|---|---|---|---|---|---|
| Scenario-1 | 232 | 35 | 78 | 2.97 (2.04) | 9 | 11.54% | ↑7.88% |
| Scenario-2 | 776 | 111 | 73 | 10.63 (4.22) | 27 | 36.99% | ↑4.9% |
| Scenario-3 | 870 | 367 | 72 | 12.08 (1.98) | 45 | 62.5% | ↑3.24% |

[1] The symbol '#' represents the number of QA pairs, entities, properties or CVTs.
[2] The compression ratio is defined as $Compr_1 = \#QA \div \#Property$ and $Compr_2 = \#QA \div (\#Entity + \#Property)$.
[3] The CVT ratio is defined as $CVTr = \#CVTProperty \div \#Property$.

We demonstrates precise question answering in Fig. 7. For the first example, given the question "优惠券和单品宝能不能一起使用 (Can coupon and SKU Bao be used at the same time?)", our bot replies with a precise and interpretable answer "Coupon is a kind of In-store Discount, SKU-Bao is a kind of SKU Discount, In-store Discount and SKU Discount can be used in conjunction". For the second example, one asks "淘抢购是否计入双十一最低价 (Whether the Tao Flash Sale is counted in the floor price of Double 11)", our bot answers with a precise "计入 (Yes)" in the concise and explicit table through identifying the "Subjsect Event", "Object Event" and"Promotion Means" slots [6], and further gives tips that help customsers to understand relevant regulations. The third example shows how the structured represenattion is used to guide customers when questions are vague: when a customer merely mention "618", our bot generates several probable questions based on the candidate properties of "Promotion_Program", the class of "618". The forth example demonstrates how our key-value structure can be used to segment long text answer and support tabbed knowledge representation.

## 5 Related Work

In this section, we review related work on knowledge representation, KBQA, and knowledge reasoning, which are closely related to the techniques employed in this paper.

**Knowledge Representation.** The key of knowledge graphs (KG) is to understand real world entities and their relationships in terms of nodes and links, i.e., things, not strings. To the best of our knowledge, most of the studies on KBQA treat knowledge graphs as a set of subject-predicate-object (SPO) triples, build their approaches on ready-made knowledge graphs such as Freebase [5], YAGO [16] and DBpedia [1], and evaluate them on benchmark data-sets（e.g., WebQuestions [3]). Few researchers have tried to apply the classic triple structure in practical scenarios. In fact, not surprisingly, this triple representation is not enough in capturing practical knowledge. We employ property chain to capture hierarchical properties, use key-value pairs to segment long text value, and adopt compound value type (CVT) from Freebase [5] to capture constraints. Our knowledge representation is evaluated in a set of realistic scenarios in the E-commerce customer service field, and it turns out to be adequate in capturing practical knowledge and supporting subsequent KBQA applications.

**KBQA.** Question answering over knowledge base, which takes as input natural language and translate it into tailor-made language (e.g., SPARQL [5]), has attracted much attention since the rise of large scale structured knowledge base such as DBpedia [1], YAGO [16], and Freebase [5].

The state-of-the-art approaches can be classified into two categories: *semantic parsing* based and *information extraction* based. Semantic parsing [3][11][4] is able to provide a deep understanding of the question, which can help to justify and interpret the answer, but it is often decoupled from knowledge base [19]. Information extraction based methods retrieve a set of candiate answers center on the topic entity in a question, and extract features from the question and candidates to rank them [6][9]. One limitation of such methods is that they are insufficient in dealing with compositional questions that involve multiple constraints.

Yih et al. [19] reduces semantic parsing into a staged query graph generation problem, uses knowledge base to prune the search space when forming the graph, and achieves a competitive result on the WebQuestions dataset. Bao et al. [2] further proposes to solve multi-constraint questions based on query graphs. On oberving that real-world questions are rather complicated, we base our work on these two approaches, and focus on constraints detection and knowledge reasoning.

**Knowledge Inference/Reasoning.** In the literature, knowledge inference is usually taken as a KB completion problem, i.e., finding missing facts in a knowledge base.

---

[6] Note that the slot "participated goods" is defaulted as "Yes".

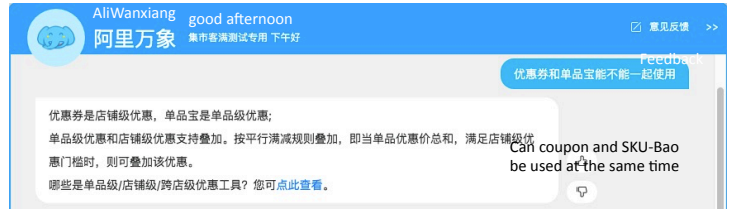

Sub–Graph (1)

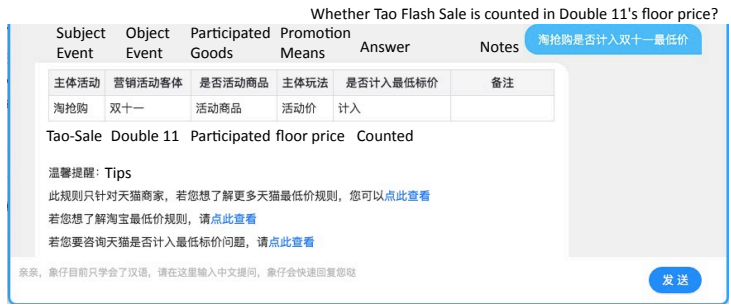

Sub–Graph (2)

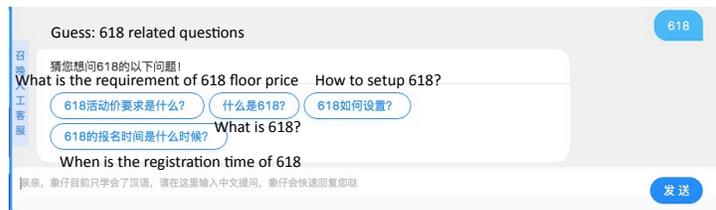

Sub–Graph (3)

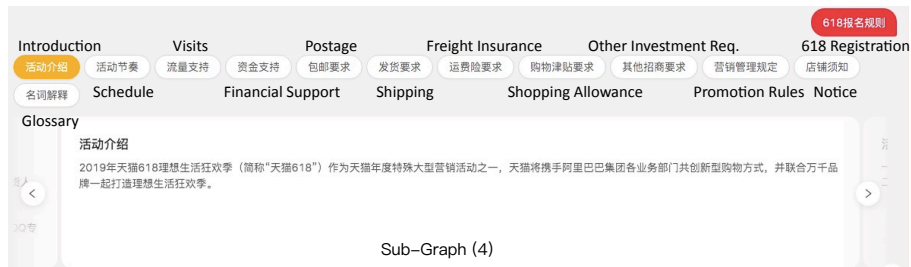

Sub–Graph (4)

**Fig. 7.** Demonstration of system features

Specifically, it includes relation inference $(e_h, ?, e_t)$ and link prediction $(e_h, p, ?)$, where $e_h, e_t \in \mathcal{E}$ and $r \in \mathcal{P}$. In general, KB reasoning methods are based on path formulas [12] or KG embeddings [7][18][14]. Our work differs from KB completion in that we perform type reasoning based on our structured knowledge representation during the question answering process.

## 6 Discussion and Conclusions

To our knowledge, this is the first attempt to apply KBQA in customer service industry. To deal with complicated knowledge and questions, we propose a novel structured knowledge representation, and accordingly introduce our approach about how to support multi-constraint and reasoning.

The benefits of our structured knowledge representation are many-fold: (1) it is able to capture class-level, rather than instance-level knowledge, hence largely reducing the number of knowledge items and making it more convenient for human management and easier for model-based text matching; (2) it defines a common structure for repetitive scenarios, with which business knowledge staffs do not need to repeatedly eliciting FAQs and training samples can be highly reused for scenarios of the same type; (3) it supports multi-constraint and reasoning, which contribute to performing precise question answering and offering better user experience; (4) it allows to guide vague or incomplete customer questions based on knowledge structure and dialog context; (5) Last but not least, it enables diversified UI representation of knowledge (e.g., key-value structured for tabbed representation, CVT for table representation), which brings about better readability and understandability, hence better customer satisfaction.

Several problems remain open. One interesting problem is how to lower the cost of manual schema construction through utlizing information extraction [15] techniques. We are also interested in investigating TableQA [17] for systematically addressing constraints recognition in our CVT knowledge structure.